%% file: iclr2023_conference_tinypaper.tex
\documentclass{article} 
\usepackage{iclr2023_conference_tinypaper,times}
\usepackage{graphicx}
\input{math_commands.tex}

\usepackage{hyperref}
\usepackage{url}

\usepackage{float}
\usepackage[utf8]{inputenc} 
\usepackage[T1]{fontenc}    
\usepackage{hyperref}       
\usepackage{url}            
\usepackage{booktabs}       
\usepackage{amsfonts}       
\usepackage{nicefrac}      
\usepackage{microtype}     
\usepackage{multirow}
\title{Solar Panel Segmentation: Self-Supervised Learning Solutions for Imperfect Datasets}

\author{
\textbf{Sankarshanaa Sagaram\textsuperscript{*}}, \textbf{Krish Didwania\textsuperscript{*}}, \textbf{Laven Srivastava\textsuperscript{*}}, 
\textbf{Aditya Kasliwal\textsuperscript{*}},\\ \textbf{Pallavi Kailas\textsuperscript{*}} \& \textbf{Ujjwal Verma\textsuperscript{\dag}} \\
\textsuperscript{*}Manipal Institute of Technology, Manipal,
\textsuperscript{\dag}Manipal Institute of Technology Bengaluru \\
Manipal Academy of Higher Education, Manipal, Udupi, India \\
\texttt{\{sanky.sagaram,krish0674,lavensri,kasliwaladitya17\}@gmail.com} \\
\texttt{kailaspallavi464@gmail.com, ujjwal.verma@manipal.edu}
}

\iclrfinalcopy 
\begin{document}

\maketitle
\

\begin{abstract}

The increasing adoption of solar energy necessitates advanced methodologies for monitoring and maintenance to ensure optimal performance of solar panel installations. A critical component in this context is the accurate segmentation of solar panels from aerial or satellite imagery, which is essential for identifying operational issues and assessing efficiency. This paper addresses the significant challenges in panel segmentation, particularly the scarcity of annotated data and the labour-intensive nature of manual annotation for supervised learning. We explore and apply Self-Supervised Learning (SSL) to solve these challenges. We demonstrate that SSL significantly enhances model generalization under various conditions and reduces dependency on manually annotated data, paving the way for robust and adaptable solar panel segmentation solutions.
\end{abstract}

\section{Introduction}

The escalating role of solar energy in mitigating climate change has garnered increased research interest, propelled by technological advancements and heightened environmental awareness as portrayed in \cite{Zhuang2020TheAS}. In this context, remote sensing has emerged as a pivotal tool for enhancing solar energy utilization, enabling the identification of regions with underutilized energy for targeted optimization (\cite{rasmussen2021challenge}). Nonetheless, segmenting solar panels with traditional supervised machine learning is challenging due to a lack of annotated datasets and the large amount of raw, unlabelled satellite data. Processing and accurately labelling this data requires extensive labour and is prone to inaccuracies, as shown by \cite{10.1145/3494832}.

Our paper addresses segmentation challenges by leveraging extensive satellite imagery and using SSL techniques, bypassing the need for meticulously annotated data as in \cite{rs14215350}. Despite label corruption, we demonstrate that SimCLR pretraining \cite{pmlr-v119-chen20j} produces accurate segmentation masks. While pretraining incurs initial costs, it is marginal compared to fully annotating datasets. Our method reduces manual labour and overall costs, presenting a cost-effective and robust solution to challenges in applying machine learning to solar energy optimization \cite{10.1093/nsr/nwx106}.

\section{Methodology}

We utilized the PV03 dataset (\cite{essd-13-5389-2021}), encompassing solar panel data from Jiangsu Province, China, and incorporated SimCLR pertaining as part of our experimental setup. Our experimentation involved various segmentation models, including encoder-decoder networks like U-Net(\cite{ronneberger2015unet}) and FPN(\cite{lin2017feature}), and PSPNet(\cite{zhao2017pyramid}), which utilizes a pyramid pooling module. Additionally, we experimented with multiple backbones for each architecture, such as ResNets(\cite{he2015deep}), Mix Visual Transformer(\cite{xie2021segformer}), and VGG(\cite{ronneberger2015unet}) with ImageNet pre-trained weights. The results of our best configurations are presented in \ref{tab:t1}. We employed Focal Loss (\cite{lin2018focal}) for all configurations, which enhances learning by prioritising challenging misclassified examples. In all our experiments, we applied horizontal and vertical flips along with colour jittering at a probability of 0.5 and shifts in the HSV colour scale. Our analysis of these configurations led us to identify the most effective settings for our approach: a batch size of 8, focal loss parameters with an $\alpha$ value of 0.4 and a $\gamma$ value of 2, a learning rate set at 3x$10^{-5}$, and the use of the Adam optimizer \cite{kingma2017adam} with $\beta_1$ and $\beta_2$ values of 0.9 and 0.99, respectively, during the fine-tuning process.

\begin{figure}[h]
  \centering
  \includegraphics[width=1\textwidth]{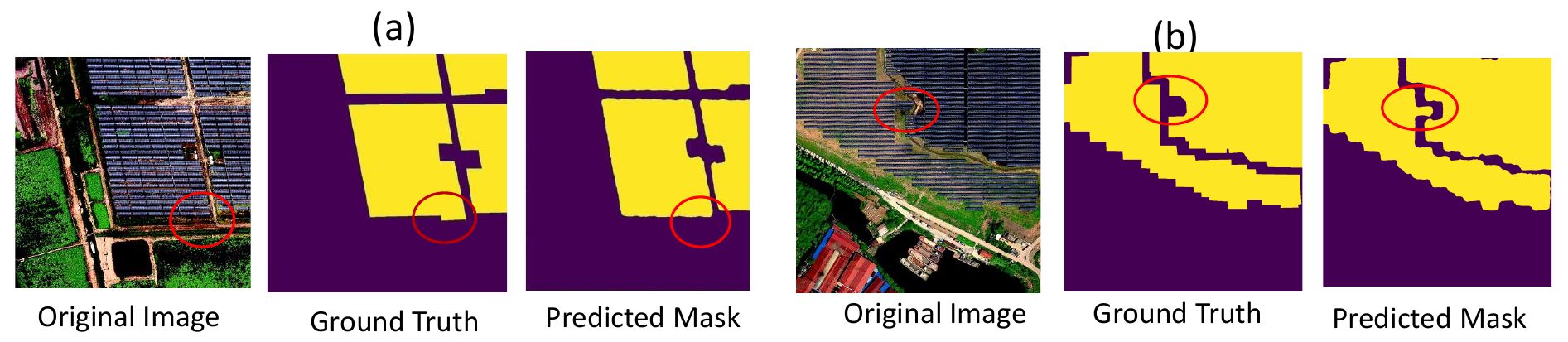}
  \caption{Instances demonstrating corrupted ground truths in the datasets, with contrast-enhanced RGB images for easier visualization, and how SSL adapts in the learning process.}
  \label{fig:figure1}
\end{figure}


\section{Results}

Initially, the entire training dataset was employed, followed by fine-tuning using varied subsets of this data before applying the model to the complete test set. Remarkably, when applied to these subsets, diverse configurations consistently produced similar results during the fine-tuning stage. Notably, it achieved impressive performance even with limited annotated data as portrayed in \cite{NEURIPS2019_a2b15837}, demonstrating that substantial annotation costs can be reduced by fine-tuning on a subset while still maintaining consistently high-quality results. Our pre-trained model was then fine-tuned on PSPNet with ResNet-34 encoder, following the methodology outlined in \cite{essd-13-5389-2021}.

\begin{table}[h]
\centering
\caption{PV03 Results with different configurations}
\label{tab:t1}
\begin{tabular}{ccccl}
\hline
\textbf{}   & \textbf{60\%} & \textbf{70\%} & \textbf{80\%} & \textbf{Full Dataset} \\ \hline
Unet [\cite{essd-13-5389-2021}]       & -             & -             & -             & 0.858                 \\
{PSPNet *} & 0.869       & 0.8841         & {0.869}        & {0.8895} \\  
{SimCLR +PSPNet *} & 0.8748        & 0.888         & \textbf{0.8911}        & {0.890} \\ \hline 
\end{tabular}
\end{table}



Our study revealed two key findings: firstly, the superior performance of our predicted masks over the original, corrupted ground truths in the dataset, and secondly, penalties in terms of potentially lower IoU scores during the fine-tuning phase of experiments that were conducted with SSL pre-training. The predicted masks accurately segmented solar panels that were missed in the manual annotation process, as clearly demonstrated in Figure \ref{fig:figure1}, where areas of interest are highlighted with red circles. This not only underscores the effectiveness of our methodology but also highlights the issue of label corruption in datasets. The lower IoU scores observed during SSL pre-training can be attributed to the generation of masks that more accurately represent real-world scenarios when compared to the provided ground truths, leading to unfair penalties during backpropagation, as seen in Figure\ref{fig:figure1}. 

\section{Conclusion}

In our research, we have effectively applied SSL techniques to tackle the challenges of scarce and corrupted data annotations in solar panel segmentation. Our approach significantly improves model generalization, even under varying conditions, thereby addressing the limitations of manual annotation in supervised learning. The experiments conducted demonstrate that our SSL-based method consistently delivers strong performance, even with limited training data, highlighting its capability to mitigate the issues of label scarcity and corrupted labels. Furthermore, the enhanced robustness achieved in this domain is a testament to the versatility of our methodology.


\subsubsection*{URM Statement}
The authors acknowledge that all key authors of this work meet the URM criteria of ICLR 2024 Tiny Papers Track.

\bibliography{iclr2023_conference_tinypaper}
\bibliographystyle{iclr2023_conference_tinypaper}

\newpage
\appendix

\section{Self-Supervised Learning and SimCLR}

Self-Supervised Learning (SSL) has emerged as a powerful paradigm in machine learning, particularly for situations where annotated data is scarce or labour-intensive to obtain. SSL leverages the data's inherent structure to generate labels, which facilitates the learning of robust feature representations without the need for extensive manual labelling. One prominent method within SSL is SimCLR (Simple Framework for Contrastive Learning of Visual Representations), which has demonstrated significant success in learning useful representations from unlabeled data through contrastive learning.

SimCLR operates by generating different augmented views of the same image and treating them as positive pairs while considering all other images in the batch as negative pairs. This process involves two key operations: attraction and repulsion. Within a batch of images, each image is augmented twice, creating two views that are intended to be close to the representation space. The positive pair (two augmented views of the same image) attract each other, while they repel all other images (negative pairs) in the batch. This mechanism ensures that the learned representations are invariant to the augmentations and distinct from other images.

The core of SimCLR is the contrastive loss function, which quantifies the similarity between positive pairs and dissimilarity between negative pairs, driving the learning process to produce discriminative features that are useful for downstream tasks. The contrastive loss function, specifically the normalized temperature-scaled cross-entropy loss (NT-Xent loss), is defined as follows:
\vspace{0.3cm}
\begin{equation}
\mathcal{L}_{i,j} = - \log \frac{\exp(\text{sim}(\mathbf{z}_i, \mathbf{z}_j) / \tau)}{\sum_{k=1}^{2N} \mathbb{1}_{[k \neq i]} \exp(\text{sim}(\mathbf{z}_i, \mathbf{z}_k) / \tau)}
\end{equation}
In this formula, \(\mathbf{z}_i\) and \(\mathbf{z}_j\) are the representations of the augmented views of the same image, \(\text{sim}(\mathbf{z}_i, \mathbf{z}_j)\) denotes the cosine similarity between \(\mathbf{z}_i\) and \(\mathbf{z}_j\), \(\tau\) is the temperature parameter, and \(N\) is the batch size. The numerator captures the similarity between the positive pair, while the denominator sums over all pairs (excluding the same view pair) to normalize the measure.

This approach has particularly benefited applications in image segmentation, object detection, and other areas where labelled data is limited. By effectively learning from the structure of the data, SimCLR and similar SSL methods provide a robust foundation for developing high-performance models with minimal reliance on annotated datasets.

\section{Loss function}

We employ the use of the {Focal Loss} \cite{lin2018focal} for the fine-tuning stages of our approach, introducing a modulating term to the cross-entropy \cite{zhang2018generalized} loss, prioritise learning on challenging misclassified examples.
The adjustable hyperparameters, $\alpha$ and $\gamma$, enable us to control the focusing effect and the rate at which the loss decreases for well-classified samples. The tuning parameter gamma governs the rate at which simple examples are down-weighted alongside alpha, which plays a pivotal role in addressing class imbalance by weighting the loss of each class.
This dynamic scaling of the cross entropy loss involves the scaling factor diminishing to zero as confidence in the correct class rises. In image segmentation tasks, the Focal loss is pivotal in handling class imbalance and accentuating challenging samples, resulting in notable improvements in segmentation performance. 
The Focal loss formula is given as follows:
\begin{equation}
\text{Focal Loss} = -\frac{1}{n} \sum_{i=1}^{n} \left( \alpha (1 - \hat{y}_i)^{\gamma} y_i \log(\hat{y}_i) + (1 - \alpha) \hat{y}_i^{\gamma} (1 - y_i) \log(1 - \hat{y}_i) \right)
\end{equation}
In the formula, $y_i$ represents the ground truth value (0 or 1) for the i-th sample, and $\hat{y}_i$ represents the corresponding predicted value from the mode. The summation runs over all n samples in the dataset.

\section{Pretraining as a method to generalize across datasets} 
\subsection{SolarDK Dataset}

We have also used another dataset, SolarDK \cite{khomiakov2022solardk}, which encompasses labelled instances from two urban municipalities in the Greater Copenhagen Region and the Danish Building Registry (BBR), to evaluate and test the generalizability of our approach across datasets. As demonstrated in the experimentation section, our approach effectively addresses the robustness and adaptability of SSL in navigating the challenges inherent in solar panel segmentation and validation.


\subsection{Generalizability}

In this section, we experimented to evaluate the robustness and generalization capabilities of our self-supervised learning (SSL) approach, specifically using SimCLR, across different datasets. We performed pre-training and fine-tuning on two distinct datasets: SolarDK and PV03. Our goal was to determine whether models trained on one dataset could effectively generalize to another, thereby demonstrating the resilience of SSL methods in the context of solar panel segmentation.

As demonstrated in Table \ref{tab:t2}, our results indicate that pre-training and fine-tuning on two different datasets produces outcomes comparable to those observed when performing both operations on the same dataset. This underscores the robustness and impressive generalization capabilities of self-supervised algorithms in this domain.

\begin{table}[h]
\caption{For cross-validation, a U-net model was used alongside the ResNet-34 backbone for SolarDK; PSPNet with the ResNet-34 was fine-tuned on the PV03 dataset.}
\vspace{1cm}
\label{tab:t2}
\centering
\begin{tabular}{ccc}
\hline
\textbf{Pre-training} & \textbf{Fine-tuning} & \textbf{Max IoU} \\ \hline
SolarDK          & SolarDK             & 0.649                \\
\textbf{PV03}                 & \textbf{SolarDK}             & \textbf{0.6971}                \\ 
PV03                 & PV03                & 0.8829                \\
\textbf{SolarDK}              & \textbf{PV03}                & \textbf{0.8928}                \\
\hline
\end{tabular}
\end{table}
\vspace{0.5cm}
The generalizability of our SSL approach indicates its potential for broad application across various datasets and scenarios. By demonstrating consistent performance across different datasets, our methodology shows promise for scalable and adaptable solutions in solar panel segmentation and beyond.

\end{document}

%% file: math_commands.tex
\usepackage{amsmath,amsfonts,bm}

\def\eqref#1{equation~\ref{#1}}

\def\1{\bm{1}}

\DeclareMathAlphabet{\mathsfit}{\encodingdefault}{\sfdefault}{m}{sl}
\SetMathAlphabet{\mathsfit}{bold}{\encodingdefault}{\sfdefault}{bx}{n}

